\documentclass[sigconf]{acmart}
\usepackage{multirow}
\usepackage{booktabs}

\newsavebox{\promptboxsave}
\newenvironment{promptbox}
  {\begin{lrbox}{\promptboxsave}%
   \begin{minipage}{\dimexpr\columnwidth-14pt\relax}}
  {\end{minipage}%
   \end{lrbox}%
   \setlength{\fboxrule}{0.5pt}%
   \setlength{\fboxsep}{6pt}%
   \fbox{\usebox{\promptboxsave}}}

\copyrightyear{2026}
\acmYear{2026}
\setcopyright{cc}
\setcctype{by}
\acmConference[JCDL '26]{The 2026 ACM/IEEE Joint Conference on Digital Libraries}{October 13--16, 2026}{Frisco, TX, USA}
\acmBooktitle{The 2026 ACM/IEEE Joint Conference on Digital Libraries (JCDL '26), October 13--16, 2026, Frisco, TX, USA}
\acmDOI{10.1145/3805696.3846040}
\acmISBN{979-8-4007-2597-5/2026/10}


\begin{document}

\title{Improving Cross-Lingual Transfer for Sequential Sentence Classification in Research Papers via Structural Similarity}
\author{Kazuhiro Yamauchi}
\orcid{0009-0000-1206-2426}
\affiliation{%
  \institution{Doshisha University}
  \city{Kyotanabe}
  \country{Japan}}
\email{yamauchi23@mm.doshisha.ac.jp}

\author{Marie Katsurai}
\orcid{0000-0003-4899-2427}
\affiliation{%
  \institution{Doshisha University}
  \city{Kyotanabe}
  \country{Japan}}
\email{katsurai@mm.doshisha.ac.jp}

\begin{abstract}
Sequential sentence classification (SSC) is an essential task for structuring scientific publications, and extending SSC research to languages other than English can improve accessibility to scientific knowledge in multilingual digital libraries. Cross-lingual transfer is a promising approach to address the scarcity of training data in non-English languages.
Prior work on other natural language processing tasks has shown the benefits of capturing linguistic similarity between source and target languages. However, SSC inherently depends on patterns at the discourse level, such as label sequences and positional regularities, which appear consistently across languages regardless of linguistic differences.
To examine the factors that determine transfer success in SSC, we constructed a multilingual SSC dataset covering 13 non-English languages collected from five academic databases.
Our cross-lingual transfer experiments, using both encoder-based and generative models, show that linguistic proximity has no consistent predictive power for transfer performance, whereas structural similarity in rhetorical organization shows a weak but consistent positive correlation across models. After controlling for source-language performance, the similarity of label distributions is the most consistent predictor.
Building on this finding, we propose a set of three methods that explicitly leverage structural information using generative models. In the in-domain evaluation, the best combination reaches parity with the strongest encoder baselines, and in transfer to languages unseen during training, it outperforms the strongest encoder baseline.
\end{abstract}


\maketitle

\section{Introduction}

Sequential sentence classification (SSC) is the task of assigning each sentence in a scientific abstract to a rhetorical role such as background, objective, method, result, or conclusion~\cite{dernoncourt-lee-2017-pubmed}. Decomposing an abstract into these roles enables retrieval based on content that goes beyond keyword matching, for example finding papers with a similar method but a different background. SSC thus serves as a foundational technology for downstream applications such as literature search, automatic summarization, and paper recommendation. Recent advancements in SSC have leveraged hierarchical architectures built on Transformer models~\cite{devlin-etal-2019-bert, jin-szolovits-2018-hierarchical, cohan-etal-2019-pretrained, brack-etal-2024-sequential} and large language models (LLMs)~\cite{lan-etal-2024-multi}, reaching unprecedented performance levels. However, these developments have been predominantly centered on English, creating a gap in the accessibility of non-English academic research.

Cross-lingual transfer learning, facilitated by multilingual pre-trained language models (mPLMs)~\cite{devlin-etal-2019-bert}, has emerged as a key strategy to bridge this gap. By transferring knowledge from a language with abundant labeled data such as English to a target language with few or no labeled abstracts, it removes the need to build a labeled dataset for each target language. The success of this transfer has long been attributed to the linguistic proximity between the source and target languages~\cite{lin-etal-2019-choosing, philippy-etal-2023-towards}. However, recent studies show this proximity to be an inconsistent predictor~\cite{blaschke-etal-2025-analyzing, desouza-etal-2024-measuring} and point instead to factors specific to the task~\cite{lin-etal-2024-mplm, yun-etal-2023-x}. This motivates identifying, for SSC, what governs cross-lingual transfer beyond linguistic distance.

In this study, we focus on the intrinsic structural properties of SSC. SSC assigns a label to each sentence, and these labels form sequences whose ordering and positions follow recurring patterns across abstracts; we refer to these patterns as an abstract's rhetorical structure. Part of this structure is shared across languages because academic abstracts follow standardized conventions typified by IMRaD (Introduction, Methods, Results, and Discussion)~\cite{sollaci-pereira-2004-imrad}, so that Background tends to appear at the beginning and Result tends to follow Method. Other parts vary by language: in our data, for instance, Russian and Chinese omit Background almost entirely and begin directly with Objective. We quantify the degree to which two languages share this structure as their structural similarity, a measure that captures both the shared conventions and the departures specific to individual languages. To test whether this property predicts transfer, we constructed a multilingual SSC dataset covering 13 non-English languages with approximately 32,000 abstracts in these languages (52,487 including the English portion of PubMed-RCT). Our empirical analysis revealed that structural similarity, defined by the alignment of label distributions, label transitions, and positional regularities, is a more consistent predictor of transfer performance than linguistic proximity.

Building on these insights, we propose a set of three methods that explicitly leverage structural information to enhance cross-lingual SSC: injecting structural knowledge into prompts, reranking candidate sequences with a trained verifier, and enforcing prediction consistency in zero-shot settings. With these methods, we show that structural information can be used not only to predict transfer but also to improve SSC directly.

The main contributions of this study are as follows:
\begin{itemize}
    \item We introduce a multilingual SSC dataset covering 13 languages and demonstrate, through empirical analysis, that structural similarity correlates more consistently with cross-lingual transfer performance than linguistic proximity.
    \item We propose a set of three methods that leverage structural information and improve macro F1 scores over existing multilingual baselines.
\end{itemize}

\section{Related Work}

\subsection{Sequential Sentence Classification}
\label{subsec:ssc}

Since the release of the PubMed-RCT benchmark~\cite{dernoncourt-lee-2017-pubmed},
SSC models have evolved from hierarchical neural
architectures~\cite{jin-szolovits-2018-hierarchical} to approaches based on
Transformers~\cite{cohan-etal-2019-pretrained}. To effectively capture sequential
dependencies and rhetorical flow, modern architectures leverage hierarchical
modeling to integrate both representations at the sentence level and context at
the document level~\cite{brack-etal-2024-sequential}. The hierarchical sequential
labeling network (HSLN)~\cite{jin-szolovits-2018-hierarchical} is representative:
Bi-LSTM encodes the tokens of each sentence and attention pooling aggregates them into a sentence vector, a
second Bi-LSTM enriches each sentence vector with context from its neighbors, a
linear layer maps the result to label scores, and a conditional random field
(CRF)~\cite{lafferty2001crf} predicts the label sequence with the highest joint
probability rather than maximizing each label independently.
\citet{brack-etal-2024-sequential} replaced the word embedding layer of HSLN with
SciBERT~\cite{beltagy-etal-2019-scibert}, an encoder pretrained on scientific
text, and evaluated the resulting model across four scientific domains. More
recently, approaches based on LLMs have shown competitive
performance~\cite{lan-etal-2024-multi}. However, existing SSC research remains
predominantly focused on English.

The CRF learns a transition matrix over labels, so that regularities such as
Method preceding Result are encoded in the model rather than left to the
classifier for each sentence. \citet{brack-etal-2024-sequential} report that when
tasks from different domains are given a common label set, sharing this output
layer across them degrades performance, and attribute this to each domain having
its own transition distribution over labels. The difference between domains thus
lies in the label sequence rather than in the sentences. We compare languages on
the same terms: Section~\ref{subsec:similarity} measures how two languages differ
in the label distributions, transitions, and positions that CRF features encode.

\subsection{Cross-Lingual Transfer Learning}
\label{subsec:xlingual}

Zero-shot cross-lingual transfer, in which a model is trained on a source
language and evaluated on a target language with no labeled examples, became
widely studied with the release of multilingual BERT
(mBERT)~\cite{devlin-etal-2019-bert}, an mPLM pretrained on 104 languages.
Transfer performance has been reported to correlate with typological
similarity~\cite{lauscher-etal-2020-zero} and alignment in word
order~\cite{deshpande-etal-2022-bert}. However, linguistic proximity is an
inconsistent predictor: its effect varies across
tasks~\cite{blaschke-etal-2025-analyzing}, and competitive transfer occurs even
without shared script or family~\cite{desouza-etal-2024-measuring}. A separate
line of work finds that similarity in the internal representations of mPLMs
tracks transfer more closely than surface features~\cite{lin-etal-2024-mplm,
yun-etal-2023-x}. These findings point to factors specific to the task, which
motivates a structural view.

Discourse-level work provides a starting point. \citet{zeyrek2020ted} annotate
parallel transcripts in six languages and report that the distribution of
discourse relations differs across them even when the content is held constant,
and \citet{braud-etal-2017-cross} report that in cross-lingual discourse parsing,
the segmentation of a document into related spans transfers reasonably well,
whereas the relations themselves degrade sharply, which they attribute in part to
differences in the relation distributions of the source and target corpora.
Comparative studies of research article abstracts report variation of the same
kind in the genre we target: English abstracts tend to argue for the significance
of the work, whereas abstracts addressed to local scientific communities tend to
report what was done, with corresponding differences in which rhetorical roles
appear and how often~\cite{martin-martin-2003-genre, vanbonn-swales-2007-english,
yakhontova-2002-selling}. These are manual analyses of small corpora, each
covering a single language pair, and they do not connect the variation they
document to automatic classification. Since SSC operates on sequences of
rhetorical roles, this variation is a candidate predictor of transfer that is
independent of linguistic proximity, and it is the one we test.

\subsection{Structure-Aware Methods}
\label{subsec:structure_aware}

Recent structure-aware approaches have improved document
modeling~\cite{buchmann-etal-2024-document} and argumentation
mining~\cite{sun-etal-2024-discourse}. The research has employed grammar
constrained decoding~\cite{geng-etal-2023-grammar, park2024grammar} and
discriminative reranking~\cite{wang-etal-2024-fantastic} to ensure output
consistency. In light of these developments, our study adapts the
verifier-reranking paradigm~\cite{cobbe2021training} to SSC, using structural
features to estimate the candidate quality without requiring gold labels. In this
paradigm, a model samples several candidate outputs and a separately trained
verifier scores each one, replacing the single greedy decode with a selection
among candidates; the verifier is trained on candidates paired with their
observed correctness, so it needs no gold labels at inference. We score whole
candidate label sequences rather than individual labels. We do not adopt
grammar-constrained decoding, since the constraint we need is distributional
rather than categorical, as Section~\ref{subsec:structural_patterns} shows.

\section{Multilingual SSC Dataset}
\label{sec:dataset}

\subsection{Data Sources and Collection}
\label{subsec:collection}

The dataset was constructed from structured scientific abstracts. Section headers supply sentence labels without manual annotation; the resulting data are intended for training models that are applied to abstracts without headers, where rhetorical roles are not directly available. Following the labeling scheme of PubMed-RCT 20k~\cite{dernoncourt-lee-2017-pubmed}, we built a multilingual dataset from non-English abstracts whose section headers provide the rhetorical labels. PubMed-RCT is derived from the MEDLINE/PubMed Baseline Database and is not released under an explicit license; we use it strictly for non-commercial research, consistent with the National Library of Medicine's terms.

We selected five academic databases based on three criteria: (1)~provision of API access to abstracts, (2)~substantial non-English content, and (3)~inclusion of structured abstracts with section headers:

\begin{itemize}
    \item \textbf{DOAJ}\footnote{\url{https://doaj.org/}}: Open-access journals across multiple languages, accessed via REST API.
    \item \textbf{HAL} (Hyper Articles en Ligne)\footnote{\url{https://hal.science/}}: the French national open archive, accessed via OAI-PMH interface.
    \item \textbf{Dialnet}\footnote{\url{https://dialnet.unirioja.es/}}: Spanish bibliographic database containing Spanish and Portuguese publications.
    \item \textbf{TRdizin}\footnote{\url{https://trdizin.gov.tr/}}: Turkish academic index, accessed via search API.
    \item \textbf{CiNii Research}\footnote{\url{https://cir.nii.ac.jp/}}: Japanese academic database, accessed via REST API.
\end{itemize}

We targeted 13 non-English languages: French, Japanese, Spanish, Chinese, Russian, Portuguese, Italian, Indonesian, Turkish, Korean, Polish, Dutch, and Estonian. Since structured abstracts typically contain Method and Result sections, we translated these two terms into each target language and used them as search queries. We used only these two terms because nearly all structured abstracts contain them and their wording varies little across journals, whereas headers for Background, Objective, and Conclusion vary widely. The queries thus condition on the presence of Method and Result headers and place no condition on the other sections. We then verified the presence of section headers through pattern matching for three formats: the XML format \texttt{<sec> <title> Methods </title> ... </sec>}, the bracket format \texttt{[Methods]}, and the colon format \texttt{Methods:}. Data collection was conducted between February and May 2025.

Headers were mapped to the five labels using keyword lists compiled for each language (for example, Aim, Purpose, and Objectives to Objective; Introduction to Background; Discussion to Conclusion). A header not in the lists was not treated as a header, and the text following it was kept in the preceding section; text preceding the first header was discarded. The lists are part of the released code.

\subsection{Preprocessing and Quality Assurance}
\label{subsec:preprocessing}

We applied the following preprocessing steps: (1)~HTML entity conversion, (2)~NFKC Unicode normalization, (3)~language detection using langdetect~\cite{shuyo2010langdetect} to discard mismatched abstracts, and (4)~duplicate removal based on title matching that ignores case. We retained only abstracts with two or more sections and used NLTK~\cite{bird-etal-2009-nltk} for sentence segmentation in European languages and spaCy~\cite{honnibal-etal-2020-spacy} for Asian languages. The headers in all three formats were removed before sentence segmentation; in the colon format, the header string was stripped from the first sentence of the section. The inputs to all models in Sections~\ref{sec:analysis} and \ref{sec:experiments} therefore contain none of the headers from which the labels were derived. PubMed-RCT likewise provides sentences without headers.

To verify dataset quality, we randomly sampled 50 abstracts from each of the nine languages with the largest amounts of data and manually inspected the labels of individual sentences. On average, 2.71 erroneous sentences were identified per 50 abstracts, which we consider negligible.

\subsection{Dataset Statistics}
\label{subsec:statistics}

The dataset contained 52,487 abstracts and 504,416 sentences across 14 languages, comprising the 13 collected languages plus English from PubMed-RCT 20k. Table~\ref{tab:dataset_stats} shows statistics by language.

Because structured abstracts are most common in medicine and life sciences, the dataset is drawn predominantly from these fields, as Table~\ref{tab:domain_dist} shows. Domain information was unavailable for Chinese, Turkish, Korean, Polish, Dutch, and Estonian from the DOAJ/TRdizin metadata; for these languages we expect a similar skew toward medicine and life sciences, since the convention of structured abstracts is largely confined to these fields. Whether the structural differences observed in Section~\ref{sec:analysis} partly reflect domain similarity is examined in Section~\ref{sec:limitations}.

The dataset and code are available on GitHub\footnote{\url{https://github.com/mm-doshisha/multilingual-SSC}}; parts of the code were written with the assistance of Anthropic's Claude and were reviewed by the authors. For data from CiNii Research, Dialnet, and TRdizin, we provide document IDs instead of full abstracts to comply with their data usage policies.

\begin{table}[t]
\centering
\small
\caption{Dataset statistics.}
\label{tab:dataset_stats}
\begin{tabular}{llrr}
\toprule
Language (abbr.)   & Source & \#Papers & \#Sentences \\ \midrule
English (en) & PubMed-RCT  & 20,000 & 180,040   \\
French (fr)      & HAL         & 11,210 & 134,393   \\
Japanese (ja)    & CiNii       & 8,366  & 78,843    \\
Spanish (es)     & Dialnet     & 5,768  & 55,743    \\
Chinese (zh)     & DOAJ        & 3,522  & 24,649    \\
Russian (ru)     & DOAJ        & 1,163  & 9,522     \\
Portuguese (pt)  & Dialnet     & 1,122  & 8,865     \\
Italian (it)     & DOAJ        & 624    & 6,353     \\
Indonesian (id)  & DOAJ        & 434    & 4,270     \\
Turkish     & TRdizin     & 179    & 630       \\
Korean       & DOAJ        & 48     & 485       \\
Polish      & DOAJ        & 30     & 369       \\
Dutch       & DOAJ        & 14     & 131       \\
Estonian    & DOAJ        & 7      & 123       \\ \bottomrule
\end{tabular}
\end{table}

\begin{table}[t]
\centering
\small
\caption{Approximate domain distribution per language; the primary domain is shown.}
\label{tab:domain_dist}
\begin{tabular}{llr}
\toprule
\textbf{Language} & \textbf{Primary domain} & \textbf{\%} \\
\midrule
English   & Medical/Life sciences (PubMed-RCT) & 100 \\
Japanese  & Medical/Life sciences (CiNii)       & 100 \\
French    & Life sciences (HAL)                   & 95.4 \\
Russian   & Medicine \& Health (DOAJ)             & 97.0 \\
Indonesian & Medicine \& Health (DOAJ)            & 73.8 \\
Italian   & Medicine \& Health (DOAJ)             & 72.2 \\
Spanish   & Health sciences (Dialnet) & 65.1 \\
Portuguese & Health sciences (Dialnet) & 63.4 \\
\bottomrule
\end{tabular}
\end{table}

\section{Cross-Lingual Transfer Analysis}
\label{sec:analysis}
For SSC, which assigns rhetorical roles at the sentence level, we hypothesize that similarity in rhetorical structure across languages predicts cross-lingual transfer more consistently than linguistic proximity. To investigate this hypothesis, we selected nine languages with 200+ abstracts from the dataset: Chinese, Spanish, English, French, Indonesian, Italian, Japanese, Portuguese, and Russian.
We conducted comprehensive cross-lingual transfer experiments across $9\times 9$ language pairs, including pairs of the same language, using both BERT and LLM-based models, examining whether the findings generalize across different model architectures.
Our experimental setting was zero-shot cross-lingual transfer: For each of the 81 language pairs, 
models were trained on the source language and evaluated on the target language 
without any training examples in the target language. 

\subsection{Models}
\label{subsec:models}

We employed mBERT with the hierarchical sequence labeling network (HSLN) architecture~\cite{brack-etal-2024-sequential}, hereafter referred to as mBERT-HSLN. HSLN uses an architecture with two layers: representations at the sentence level are first obtained by aggregating token embeddings via a Bi-LSTM, and then a second Bi-LSTM performs contextualized sequence labeling over the sentence representations. We followed the hyperparameters from \citet{brack-etal-2024-sequential}.

For LLM-based models, we selected three models fine-tuned to follow natural language instructions: Gemma2-2B-it~\cite{gemma2team2024}, Qwen2.5-3B-Instruct~\cite{qwen25team2024}, and Llama-3.2-3B-Instruct~\cite{llama3team2024}. The prompt used was a simplified version of that proposed by \citet{lan-etal-2024-multi}, from which we removed the demonstration examples (to focus on zero-shot transfer) and the abstract context (for the reason given below). We unified prompts in English for all experiments in this section. Each prompt contains only the sentence to be classified, without the surrounding sentences; each sentence is classified in a separate call, and the label sequence of an abstract is the concatenation of these outputs. We chose this setting for the transfer analysis so that the LLMs cannot exploit the order and position of sentences, for which they could rely on knowledge acquired in pretraining rather than on the source corpus, whereas HSLN models the sequence only through components trained from scratch on that corpus. The template is shown in Figure~\ref{fig:prompt_base}.

\begin{figure}[t]
\centering
\begin{promptbox}
\small
\textbf{Instruction:}  'You must categorize the given sentence into one of these five labels: Background, Objective, Method, Result, Conclusion. Respond with ONLY the label name.'

\textbf{Input:} 'Target Sentence: Individuals who received CM targeting psycho-stimulants were 79\% more likely to submit a smoking-negative breath-sample relative to controls.'

\textbf{Output:} 'Question: What is the rhetorical role of the Target Sentence? Answer with one word from the labels list.'
\end{promptbox}
\caption{Prompt template used in the transfer analysis. The example sentence is taken from the training split of PubMed-RCT 20k.}
\label{fig:prompt_base}
\end{figure}

We applied low-rank adaptation (LoRA;~\cite{hu-etal-2022-lora}) fine-tuning with $r=32$ and $\alpha=64$ for all attention and feedforward network layers, using a learning rate of $2\times10^{-4}$, batch size of 4, and training for three epochs.

For each language, we split the data as follows: 70\% training, 15\% validation, and 15\% testing. The test set for each language comprised $\min(200,\ \lfloor 0.15 \times N \rfloor)$ abstracts, where $N$ is the total number of abstracts for that language. For each language pair, we conducted three runs with different random seeds and recorded the average Macro~F1 scores.

\subsection{Results of Zero-Shot Cross-Lingual Transfer}

\begin{figure}[t]
  \centering
  \includegraphics[width=\columnwidth]{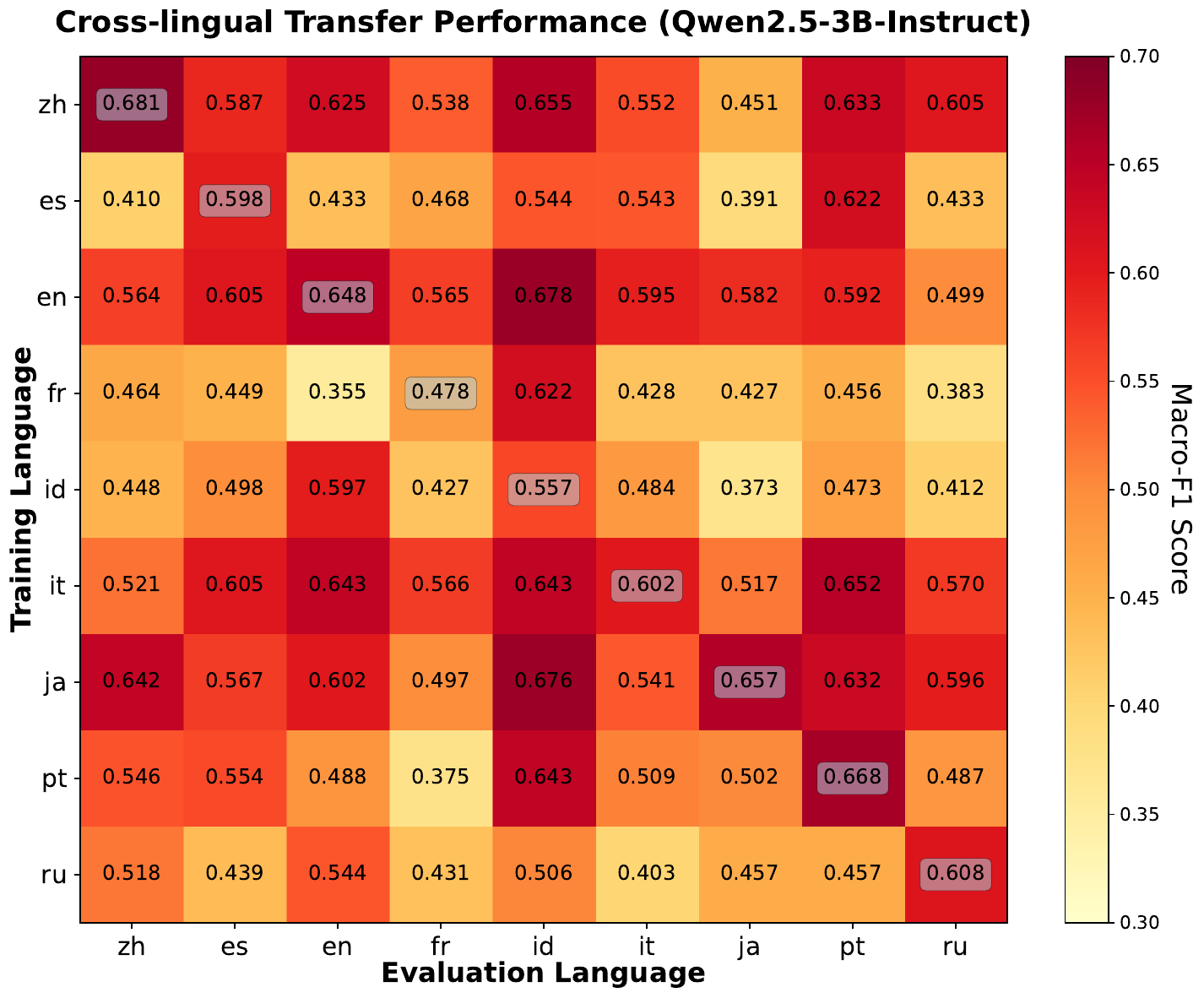}
  \caption{Cross-lingual transfer performance for Qwen2.5-3B-Instruct (Macro~F1). Rows: training languages; Columns: evaluation languages.}
  \label{fig:transfer_heatmap}
\end{figure}

Figure~\ref{fig:transfer_heatmap} shows the transfer results for Qwen2.5-3B-Instruct as a representative example. We observed consistent transfer patterns across all four models, namely mBERT-HSLN and the three LLMs, which suggests that the findings are not specific to particular model architectures. Pairs of the same language reached an average Macro~F1 of 0.611 for this model, while cross-lingual transfer averaged 0.525. Transfer performance varied considerably: Japanese to Chinese and Chinese to Indonesian reached 0.642 and 0.655, whereas French to English and Spanish to Japanese fell to 0.355 and 0.391. For mBERT-HSLN, the corresponding averages were 0.488 for pairs of the same language and 0.394 for cross-lingual pairs, and the strongest and weakest pairs were largely the same across architectures.

Common patterns held across all models, including stable performance with English as source. Transfer was also asymmetric; for example, Qwen transferred Japanese to Chinese at 0.642 but Chinese to Japanese at only 0.451, likely reflecting model capability differences for each language. Notably, Japanese and English consistently achieved high performance as source languages across all four models, while French and Russian were among the weakest sources. This asymmetry is consistent with the source language effect discussed in Section~\ref{subsec:confounding}, where transfer performance correlates with the source language's in-domain performance.

\subsection{Linguistic and Structural Similarity Measures}
\label{subsec:similarity}
To determine which factors predict cross-lingual transfer performance, we compared two similarity measures between language pairs: linguistic proximity and structural similarity.
For linguistic proximity, we used lang2vec~\cite{littell2017uriel}, which provides typological feature vectors for languages. We concatenated feature vectors from five categories: syntax, phonology, inventory, geography, and family, and calculated the cosine similarity between each language pair.

For structural similarity, we designed measures based on the rhetorical structure patterns found in the abstracts. These measures are computed on the corpus collected for each language, so the similarity between two languages is measured on these corpora and may partly reflect their source databases and disciplines rather than the languages alone (Section~\ref{sec:limitations}). Drawing on feature representations used in conditional random fields (CRF;~\cite{lafferty2001crf}) and research on abstract composition patterns~\cite{martin-martin-2003-genre}, we defined six distance measures between language pairs, all normalized to the $[0, 1]$ range. Label distribution distance uses the Jensen-Shannon divergence (JSD) between the frequency distributions of the five labels:
\begin{equation}
d_{\text{label}}(L_1, L_2) = \text{JSD}(P_{L_1}, P_{L_2})
\end{equation}
where $P_L(l)$ is the frequency of label $l$ in language $L$, computed as the number of sentences labeled $l$ divided by the total number of sentences in $L$. Section length distance averages the JSD of section length distributions across labels, where a section length is the number of consecutive sentences with the same label:
\begin{equation}
d_{\text{section}}(L_1, L_2) = \frac{1}{5}\sum_{l} \text{JSD}(P_{L_1,l}, P_{L_2,l})
\end{equation}
Continuation probability distance uses the normalized Euclidean distance between vectors of label continuation probabilities $p_L(l) = P(y_{i+1} = l \mid y_i = l)$:
\begin{equation}
d_{\text{cont}}(L_1, L_2) = \frac{1}{\sqrt{5}}\sqrt{\sum_{l}(p_{L_1}(l) - p_{L_2}(l))^2}
\end{equation}
Boundary position distance computes the weighted difference in average relative positions of four major transitions such as Method $\to$ Result, with weights proportional to the minimum transition frequency. Block count distance applies JSD to the distribution of label spans that are not contiguous, analogously to section length distance. Transition entropy distance normalizes the absolute difference in Shannon entropy of label transition distributions by $\log_2 25$, the maximum entropy for 25 transition patterns.

The overall structural distance is the simple average of all six normalized distances, $d(L_1, L_2) = \frac{1}{6}\sum_{i=1}^{6} d_i(L_1, L_2)$, and structural similarity is defined as $1 - d(L_1, L_2)$. Equal weighting was adopted as a neutral baseline to avoid overfitting given the limited sample size of 72 language pairs; weighting learned or adapted to the task was left for future work.

\begin{figure}[t]
  \centering
  \includegraphics[width=\columnwidth]{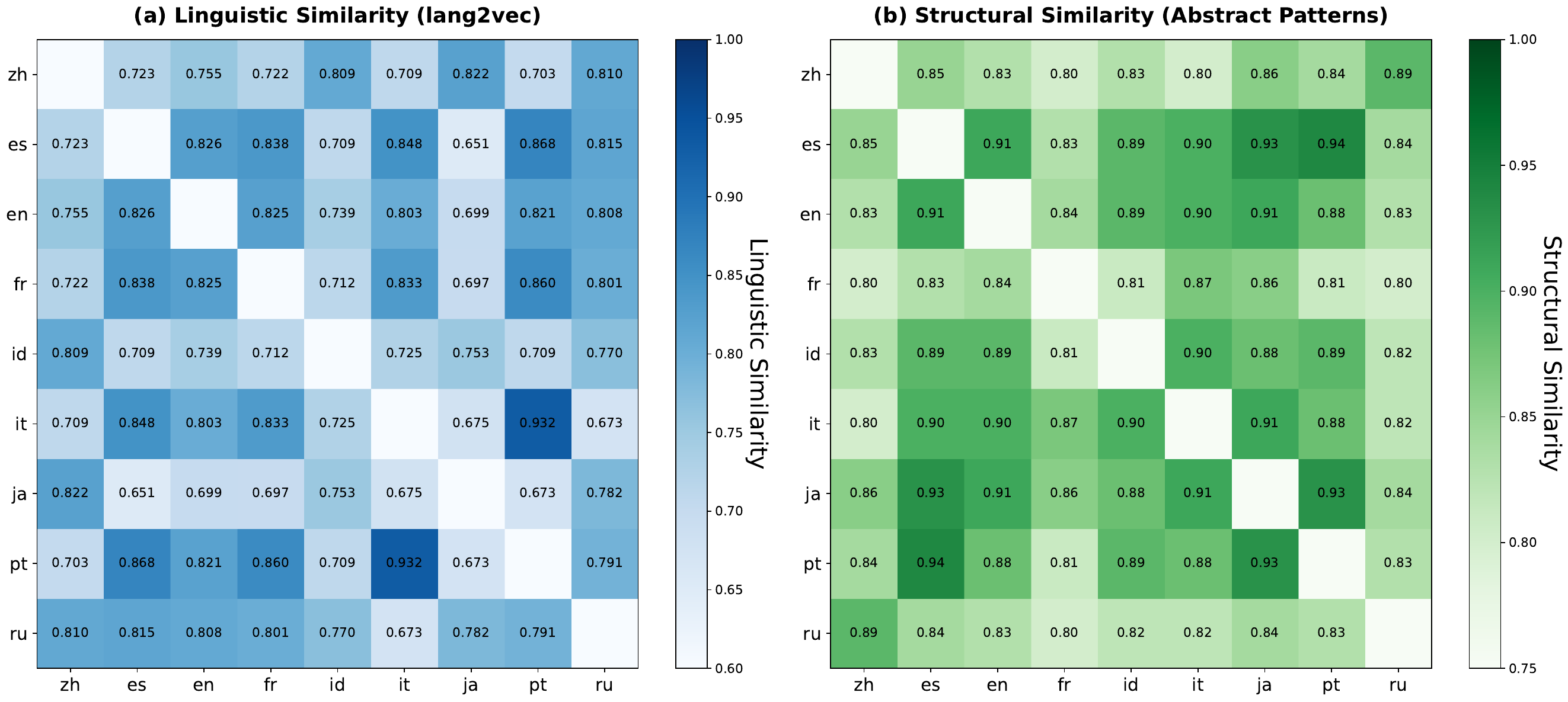}
  \caption{Linguistic (left) and structural (right) similarity matrices.}
  \label{fig:similarity}
\end{figure}

Figure~\ref{fig:similarity} shows the two similarity matrices. While linguistic proximity largely reflects language family membership, structural similarity reveals different patterns: Japanese--Spanish and Japanese--Portuguese both reach 0.93 despite belonging to different language families. Chinese--Russian also shows high structural similarity at 0.89, as both tend to omit Background and start with Objective.

\subsection{Structural Patterns Across Languages}
\label{subsec:structural_patterns}

To understand why structural similarity diverges from linguistic proximity, we analyzed the rhetorical patterns of each language using label transition probabilities and positional distributions. Two primary patterns emerged, with additional variation in positional distributions.

\textbf{Background omission.} Russian abstracts contained no Background section, and Chinese abstracts contained only four Background sentences out of 24,649. Both languages show strong Objective $\to$ Method transitions, at 0.96 for Russian and 0.95 for Chinese, indicating that abstracts begin directly with Objectives. This shared convention explains the high structural similarity of 0.89 and the above-average transfer from Chinese to Russian (0.605 for Qwen) despite their typological distance. Among the nine analyzed languages, only Chinese and Russian exhibit this pattern, forming a distinct structural cluster.

\textbf{Canonical IMRaD adherence.} Indonesian and Italian show strong sequential transitions consistent with the canonical IMRaD structure; Indonesian reaches 0.73 for Method$\to$Result and 0.63 for Result$\to$Conclusion, and Italian reaches 0.72 and 0.79 for these same transitions, with Background appearing in the first 20\% of abstracts, followed by Objective, Method, Result, and Conclusion in sequence. Similarly, Spanish, Portuguese, and Japanese abstracts maintain clear positional separation of sections, which accounts for the high structural similarity among these five languages, all pairwise similarities being $\geq 0.91$. This cluster also includes English, whose pairwise similarity with all five is $\geq 0.86$, reflecting the influence of international publication norms in biomedicine.

\textbf{Positional variation.} Beyond these two patterns, languages differ in where sections are positioned within the abstract. Because Chinese and Russian omit Background, they concentrate Objective and Method in the first 30\% of the abstract, whereas Indonesian and Italian show a more evenly distributed positional profile. French occupies an intermediate position: it retains Background sections, but the positional distributions are more compressed than those of Spanish or Portuguese, potentially reflecting differences in editorial conventions across journals in French. These positional differences are captured by the boundary position and label distribution metrics defined in Section~\ref{subsec:similarity}.

These patterns connect to cross-lingual transfer. Pairs that share rhetorical conventions despite typological distance, such as Chinese and Russian, which both omit Background, or Japanese together with Spanish and Portuguese, which all maintain clear positional separation, show high structural similarity, and across the 72 language pairs structural similarity correlates with transfer performance, as shown in Section~\ref{subsec:confounding}. At the level of individual pairs, transfer is also shaped by source language capability and is often asymmetric, so structural similarity alone does not determine any single outcome; structurally divergent pairs such as French$\to$English at 0.355 transfer poorly. The structural conventions themselves likely reflect a combination of disciplinary norms, such as medical journals requiring structured abstracts, and editorial practices specific to each language.

\subsection{Correlation Analysis}
\label{subsec:confounding}

\begin{figure}[t]
  \centering
  \includegraphics[width=\columnwidth]{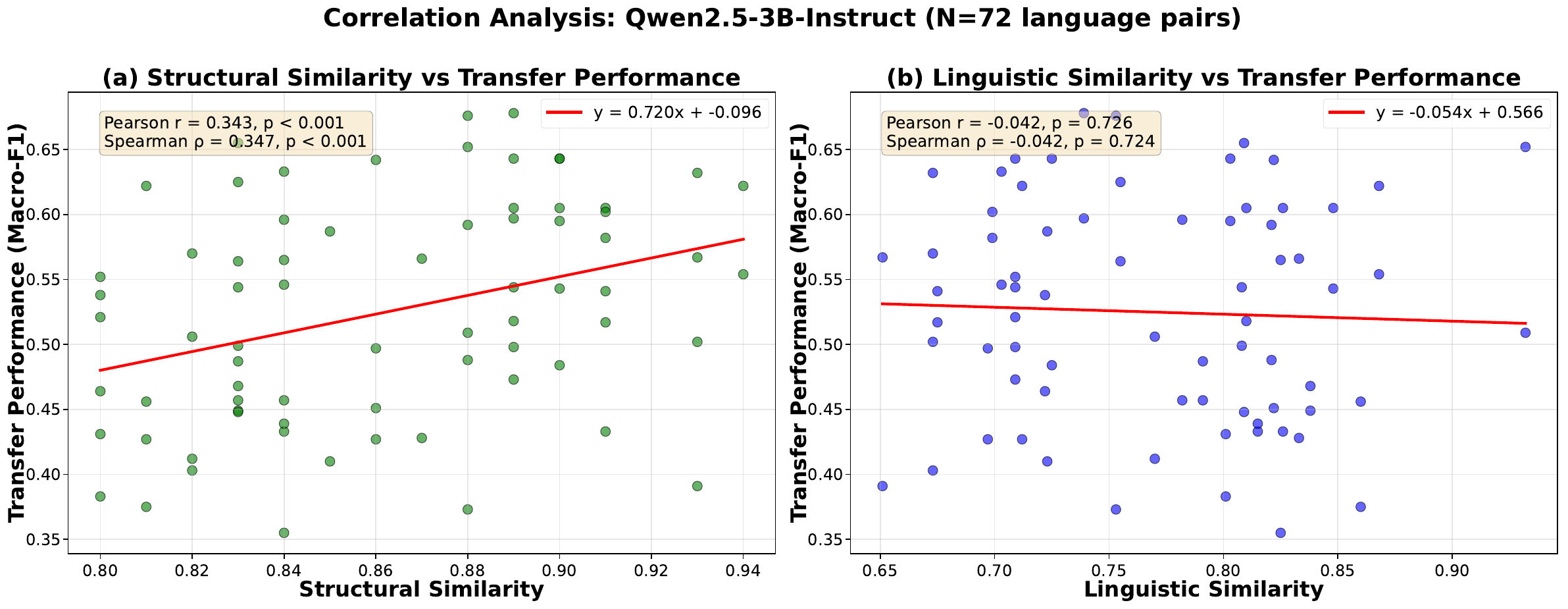}
  \caption{Correlation between similarity measures and transfer performance for Qwen2.5-3B ($N=72$). Left: structural similarity (Pearson $r=0.343$, $p<0.001$). Right: linguistic proximity (Pearson $r=-0.042$, $p=0.726$).}
  \label{fig:correlation}
\end{figure}

To determine which similarity measure better predicts transfer performance, we analyzed the 72 pairs, excluding pairs of the same language, through correlation analysis. Figure~\ref{fig:correlation} shows the results. Statistically significant positive Pearson correlations between structural similarity and transfer performance were found for all four models: mBERT-HSLN ($r = 0.372$, $p < 0.001$), Qwen2.5-3B ($r = 0.343$, $p < 0.001$), Gemma2-2B ($r = 0.308$, $p < 0.01$), and Llama-3.2-3B ($r = 0.312$, $p < 0.01$). Spearman correlations showed similar tendencies. The p-values here and in Tables~\ref{tab:metric_ablation} and \ref{tab:metric_partial} treat the 72 pairs as independent; pairs that share a language are not, so these values are nominal and likely optimistic (Section~\ref{sec:limitations}).

For linguistic proximity, no model presented consistent significant correlations. In mBERT-HSLN, neither the Pearson nor the Spearman correlation was significant ($r = 0.087$, $p = 0.438$; $\rho = 0.024$, $p = 0.831$). Among the LLMs, only Gemma2-2B showed a significant Pearson correlation ($r = 0.278$, $p = 0.012$), but the Spearman correlation was not significant. In Qwen2.5-3B and Llama-3.2-3B, no significant correlations were found (all $p > 0.05$). Overall, linguistic proximity showed weak and inconsistent predictive power.

However, transfer performance is also strongly correlated with the source language's in-domain performance (mBERT-HSLN: $r = 0.86$; Qwen: $r = 0.49$; Gemma: $r = 0.71$; Llama: $r = 0.72$; all $p < 0.001$), which is a potential confound for the correlations reported above.

\textbf{Partial correlation analysis.} To assess the independent contribution of structural similarity, we computed partial correlations controlling for the source language's in-domain performance. The combined structural similarity metric retained a significant partial correlation for Qwen ($r = 0.274$, $p = 0.021$) and Llama ($r = 0.269$, $p = 0.024$), but not for mBERT-HSLN ($r = 0.153$, $p = 0.204$) or Gemma ($r = 0.080$, $p = 0.506$). Linguistic proximity showed no significant partial correlation for any model (all $p > 0.45$). This indicates that while the overall structural similarity signal is partially attributable to source language performance, it retains independent predictive power in some architectures, whereas linguistic proximity does not.

\textbf{Individual metric contributions.} Before controlling for the confound, label distribution was significant across all four architectures, with $r$ ranging from 0.288 to 0.432. The full Pearson correlations for each metric across the four models are reported in Table~\ref{tab:metric_ablation}.

\begin{table}[t]
\centering
\footnotesize
\setlength{\tabcolsep}{3pt}
\caption{Pearson $r$ between individual structural metrics and transfer performance across four models. $^{***}p<0.001$, $^{**}p<0.01$, $^{*}p<0.05$, n.s.\ = not significant.}
\label{tab:metric_ablation}
\begin{tabular}{lcccc}
\toprule
\textbf{Metric} & \textbf{mBERT} & \textbf{Qwen} & \textbf{Gemma} & \textbf{Llama} \\
\midrule
Label distrib.     & 0.432$^{***}$ & 0.418$^{***}$ & 0.288$^{**}$ & 0.335$^{**}$ \\
Section length         & 0.234$^{*}$ & 0.245$^{*}$ & 0.241$^{*}$ & 0.223$^{*}$ \\
Contin.\ prob.     & 0.383$^{***}$ & 0.430$^{***}$ & 0.213$^{\text{n.s.}}$ & 0.184$^{\text{n.s.}}$ \\
Boundary pos.      & 0.215$^{\text{n.s.}}$ & 0.412$^{***}$ & 0.365$^{***}$ & 0.126$^{\text{n.s.}}$ \\
Block count            & 0.354$^{**}$ & 0.339$^{**}$ & 0.394$^{***}$ & 0.277$^{*}$ \\
Trans.\ entropy     & 0.354$^{**}$ & 0.388$^{***}$ & 0.198$^{\text{n.s.}}$ & 0.318$^{**}$ \\
\midrule
\textbf{Combined} & \textbf{0.372}$^{***}$ & \textbf{0.343}$^{***}$ & \textbf{0.308}$^{**}$ & \textbf{0.312}$^{**}$ \\
\bottomrule
\end{tabular}
\end{table}

\textbf{Partial correlations of individual metrics.} To identify which structural aspects retain predictive power after controlling for source language performance, we computed partial correlations for each individual metric, reported in Table~\ref{tab:metric_partial}. Label distribution was the only individual metric to remain significant in three of four models (mBERT: $r = 0.347$, $p < 0.01$; Qwen: $r = 0.251$, $p < 0.05$; Llama: $r = 0.406$, $p < 0.001$). Block count and transition entropy each reached significance in two models, whereas boundary position lost all significance, indicating that its apparent predictive power was largely attributable to source language performance. The variation across models, with Gemma showing no significant partial correlations for any metric, likely reflects differences in multilingual capability and pretraining data composition. We also note that for Qwen, the strongest model in the in-domain evaluation reported in Table~\ref{tab:indomain_overall}, the highest individual partial correlation was transition entropy at 0.337 rather than label distribution; we treat this as a pattern specific to the model and do not read it as an explanation of Qwen's overall performance. The prominence of label distribution does not stem from the sentence-level prompt of the LLMs (Section~\ref{subsec:models}): mBERT-HSLN, which uses the sequence, shows it as well.

\begin{table}[t]
\centering
\footnotesize
\setlength{\tabcolsep}{3pt}
\caption{Partial Pearson $r$ between individual structural metrics and transfer performance, controlling for source language in-domain performance. $^{***}p<0.001$, $^{**}p<0.01$, $^{*}p<0.05$.}
\label{tab:metric_partial}
\begin{tabular}{lcccc}
\toprule
\textbf{Metric} & \textbf{mBERT} & \textbf{Qwen} & \textbf{Gemma} & \textbf{Llama} \\
\midrule
Label distrib.  & 0.347$^{**}$ & 0.251$^{*}$ & 0.146 & 0.406$^{***}$ \\
Section length  & 0.029 & 0.296$^{*}$ & 0.049 & 0.180 \\
Contin.\ prob.  & 0.078 & 0.298$^{*}$ & $-0.010$ & 0.142 \\
Boundary pos.   & 0.088 & 0.105 & $-0.028$ & 0.110 \\
Block count     & 0.311$^{**}$ & 0.185 & 0.047 & 0.320$^{**}$ \\
Trans.\ entropy & 0.098 & 0.337$^{**}$ & 0.159 & 0.269$^{*}$ \\
\bottomrule
\end{tabular}
\end{table}

Together, these results indicate that structural similarity is a more relevant predictor of cross-lingual SSC transfer than linguistic proximity. The combined metric retains an independent contribution in two of four models; label distribution similarity is the individual metric that most consistently retains significance, whereas boundary position retains none.

\section{Leveraging Structural Information}
\label{sec:methods}

Section~\ref{sec:analysis} shows that the rhetorical structure of abstracts is associated with cross-lingual transfer, which motivates giving models explicit structural guidance. We therefore developed a set of three methods for explicitly leveraging structural information: structure-informed prompting (SIP), structure-guided verifier reranking (SGVR), and structure-adaptive verifier (SAV).

Unlike the structure-aware approaches in Section~\ref{subsec:structure_aware}, our methods use the structural features of the abstract itself, not the similarity between languages measured in Section~\ref{sec:analysis}, as cues at the input stage and during candidate selection, and require no gold labels at inference time.

\subsection{SIP}

SIP guides LLM predictions by explicitly incorporating structural constraints into prompts. Unlike the prompt in Section~\ref{sec:analysis}, SIP provides the full abstract together with the target sentence, and it specifies the typical label ordering of Background, Objective, Method, Result, and Conclusion so that the model can relate the position of the target sentence within the abstract to this ordering. The prompt consisted of: (1)~a task description with label definitions, (2)~structural constraints indicating typical ordering, (3)~one demonstration example randomly selected from the training data, independently for each test abstract, and (4)~the full abstract and the target sentence. The template is shown in Figure~\ref{fig:prompt_sip}.

\begin{figure}[t]
\centering
\begin{promptbox}
\small
\textbf{Instruction:} You must categorize the given sentence into one of these five labels: Background, Objective, Method, Result, Conclusion. Respond with ONLY the label name.

\textbf{Structural Constraints:} Academic abstracts typically follow this order: Background (introducing the topic) $\to$ Objective (stating the research goal) $\to$ Method (describing the approach) $\to$ Result (presenting findings) $\to$ Conclusion (summarizing implications).

\textbf{Example:} One abstract from the training data with each sentence followed by its label, selected independently for each test abstract.

\textbf{Input:} 'Abstract: Treatments for drug addiction and smoking in severely mentally ill (SMI) adults are needed. To investigate the effect of a contingency management (CM) intervention targeting psycho-stimulant on cigarette smoking. 126 stimulant dependent SMI smokers were assigned to CM or a non-contingent control condition. Rates of smoking-negative ($<$3 ppm) carbon monoxide breath-samples were compared. Individuals who received CM targeting psycho-stimulants were 79\% more likely to submit a smoking-negative breath-sample relative to controls. This study provides initial evidence that a behavioral treatment for drug use results in reductions in cigarette smoking in SMI adults.' 'Target Sentence: Individuals who received CM targeting psycho-stimulants were 79\% more likely to submit a smoking-negative breath-sample relative to controls.'

\textbf{Output:} Question: What is the rhetorical role of the Target Sentence? Answer with one word from the labels list.
\end{promptbox}
\caption{Prompt template for SIP. The abstract in the Input field is taken from the training split of PubMed-RCT 20k.}
\label{fig:prompt_sip}
\end{figure}

\subsection{SGVR}

SGVR reranks multiple candidates generated by an LLM via temperature sampling, using structural features. We adapt the verifier-reranking paradigm~\cite{cobbe2021training} to SSC: following quality estimation in machine translation~\cite{specia2018quality}, the verifier predicts output quality without reference labels and selects the most valid candidate.

Specifically, for each training abstract, the LLM generated $K$ candidate label sequences $\mathbf{y} = (y_1, \ldots, y_n)$, where $n$ is the number of sentences in the abstract and $y_i$ is the predicted label for the $i$-th sentence. Each candidate was obtained by sampling one label per sentence with temperature sampling and concatenating the samples in sentence order. We computed the accuracy at the sentence level for each candidate using the formula $q = \frac{1}{n}\sum_{i=1}^{n} \mathbf{1}[y_i = y_i^*]$, where $y_i^*$ denotes the ground truth label. From each candidate, we also extracted a feature vector of 44 dimensions $\mathbf{f}(\mathbf{y})$, listed in Table~\ref{tab:sgvr_features}, capturing (1)~label distribution features, namely occurrence counts and cosine similarity with the training data distribution; (2)~transition features, namely the mean log probability of adjacent label transitions, binary indicators of major transitions, and entropy; and (3)~position features, namely scores evaluating whether label positions fall within expected ranges and one-hot representations of start and end labels. We trained a regression model $V_\phi$ that predicts $q$ from structural features $\mathbf{f}(\mathbf{y})$.

\begin{table}[t]
\centering
\small
\caption{Structural features for SGVR, totaling 44 dimensions.}
\label{tab:sgvr_features}
\setlength{\tabcolsep}{4pt}
\begin{tabular}{lp{3.8cm}r}
\toprule
\textbf{Group} & \textbf{Feature} & \textbf{Dim} \\
\midrule
Basic scores & Transition, position, distribution & 3 \\
Sequence & Length, label occurrence counts & 6 \\
Sections & Mean, std of section length/label & 10 \\
Continuation & Continuation prob.\ per label & 5 \\
Blocks & Number of blocks per label & 5 \\
Transitions & Presence of major transitions & 4 \\
Entropy & Transition entropy & 1 \\
Boundary & Start/end labels (one-hot) & 10 \\
\midrule
\textbf{Total} & & \textbf{44} \\
\bottomrule
\end{tabular}
\end{table}

At inference, the verifier predicts accuracy $\hat{q}_k = V_\phi(\mathbf{f}(\mathbf{y}_k))$ for each candidate and selects $\mathbf{y}^{\text{pred}} = \arg\max_{k} \hat{q}_k$.

\subsection{SAV}

SAV is an unsupervised method for zero-shot settings, where no validation data are available for verifier training. Instead, it dynamically estimates structural properties from the candidate set and favors labels that are consistent across candidates. Simple majority voting, such as self-consistency~\cite{wang2023selfconsistency}, is ineffective for sequence labeling if exact matches across full sequences are rare. SAV instead evaluates agreement at each position and transition consistency separately to select the most structurally valid candidate.

For each abstract, as in SGVR, an LLM generated $K$ candidate predictions. We denote the $k$-th candidate as $\mathbf{y}^{(k)} = (y_1^{(k)}, \ldots, y_n^{(k)})$. Each candidate was evaluated using three scores:

The \textbf{consistency score} $S_{\text{cons}}(\mathbf{y}^{(k)})$ measures the average agreement rate at each position between candidate $\mathbf{y}^{(k)}$ and all other candidates. The \textbf{confidence score} $S_{\text{conf}}(\mathbf{y}^{(k)})$ first identifies the most frequent label $l_i^*$ at each position $i$ across the $K$ candidates and its agreement rate $r_i = \frac{1}{K}\sum_{j=1}^{K}\mathbf{1}[y_i^{(j)} = l_i^*]$; the score rewards candidates that agree with these majority labels at each position, weighted by the agreement rates. The \textbf{transition score} $S_{\text{trans}}(\mathbf{y}^{(k)})$ applies the same principle to transitions between adjacent labels rather than individual labels.

The final score is $S(\mathbf{y}^{(k)}) = \frac{1}{3}(S_{\text{cons}} + S_{\text{conf}} + S_{\text{trans}})$, and the prediction is $\mathbf{y}^{\text{pred}} = \arg\max_{k} S(\mathbf{y}^{(k)})$. Unlike simple majority voting at each position, SAV always outputs a sequence that exists in the candidate set, avoiding inconsistent reconstructions.

Together, SIP, SGVR, and SAV incorporate structural constraints at the input stage as well as during candidate selection. SIP+SGVR was used when the target language is included in the training data, enabling verifier training, and SIP+SAV was used for zero-shot settings.

\section{Experiments}
\label{sec:experiments}

\subsection{Experimental Settings}

Unlike the analysis of language pairs in Section~\ref{sec:analysis}, we trained models on mixed multilingual data and evaluated them in two settings: (1)~in-domain evaluation, testing on trained languages, and (2)~zero-shot evaluation, testing on languages not included during training.

For in-domain evaluation, we used nine languages: English, French, Japanese, Spanish, Chinese, Indonesian, Portuguese, Italian, and Russian, splitting each into a 70:15:15 ratio for training, validation, and testing, respectively. The model was trained on combined training data from all nine languages. For zero-shot evaluation, we used five languages not included during training: Estonian, Korean, Dutch, Polish, and Turkish. Since these languages were not used for training, their entire sets served as test sets.

We clarify how the single demonstration in SIP relates to our zero-shot claim, since the two operate at different levels. \emph{Zero-shot cross-lingual transfer} refers to the absence of any training data in the target language, whereas the single demonstration in SIP is a form of \emph{1-shot prompting} that conditions the model with one labeled example. In the zero-shot evaluation, this demonstration is drawn from the in-domain training languages, so the target language remains unseen during both fine-tuning and prompting.

We applied LoRA with $r=32$ and $\alpha=64$ for all attention and feedforward network layers to Qwen2.5-3B-Instruct, Llama-3.2-3B-Instruct, and Gemma2-2B-it, training with a learning rate of $2\times10^{-4}$ and batch size 4, for three epochs. For SGVR, we set $K=3$; for SAV in zero-shot evaluation, we set $K=5$ to balance candidate diversity against computational cost. In preliminary experiments, performance gains plateaued beyond $K=5$, while inference time increased roughly linearly with $K$. The SGVR verifier $V_\phi$ is a LightGBM~\cite{ke2017lightgbm} regressor, chosen for its ability to capture non-linear feature interactions. SGVR with $K=3$ and SAV with $K=5$ increase inference time by factors of approximately three and five relative to decoding a single candidate, and the SIP prompt is longer than the baseline prompt of Table~\ref{tab:ablation} by the ordering constraint and the demonstration. On an H100 GPU, SGVR with $K=3$ raised inference time from 1.08 to 3.35 seconds per abstract, and SAV with $K=5$ from 1.33 to 6.55 seconds. For applications sensitive to latency, SIP alone offers a competitive alternative (Table~\ref{tab:indomain_overall}). All experiments were run with three inference iterations, and we report mean values with standard deviations.

We used the following baselines: (1)~\textbf{multilingual LLM-SSC}, a multilingual extension of \citet{lan-etal-2024-multi} with their original prompt, which includes the full abstract and demonstration examples; (2) ~\textbf{mB ERT-HSLN}, the HSLN used by \citet{brack-etal-2024-sequential} with mBERT encoder; and additionally (3)~\textbf{mmBERT-HSLN}, \textbf{XLM-R-HSLN}~\cite{conneau-etal-2020-unsupervised}, and \textbf{LaBSE-HSLN}~\cite{feng-etal-2022-language}, which replace the encoder component with larger or more recent multilingual encoders. We report accuracy at the sentence level and Macro~F1 across five classes. Overall scores in Tables~\ref{tab:indomain_overall}, \ref{tab:perlang_main}, and \ref{tab:zeroshot_overall} are computed over the pooled test sentences of all languages.

\subsection{In-Domain Evaluation}

\begin{table}[t]
\centering
\small
\caption{Overall in-domain evaluation results on 9 languages.}
\label{tab:indomain_overall}
\begin{tabular}{lcc}
\toprule
\textbf{Method} & \textbf{Accuracy} & \textbf{Macro F1} \\
\midrule
mBERT-HSLN & 90.6$_{\pm0.5}$ & 82.6$_{\pm0.7}$ \\
mmBERT-HSLN & 92.2$_{\pm0.5}$ & 84.6$_{\pm0.3}$ \\
XLM-R-HSLN & 92.0$_{\pm0.9}$ & 83.9$_{\pm1.5}$ \\
LaBSE-HSLN & 83.6$_{\pm0.2}$ & 75.8$_{\pm0.1}$ \\
\midrule
LLM-SSC (Gemma) & 66.6$_{\pm0.8}$ & 57.1$_{\pm1.1}$ \\
SIP+SGVR (Gemma) & 85.5$_{\pm0.1}$ & 77.5$_{\pm0.1}$ \\
\midrule
LLM-SSC (Llama) & 63.3$_{\pm0.3}$ & 54.4$_{\pm0.4}$ \\
SIP+SGVR (Llama) & 87.5$_{\pm0.1}$ & 79.1$_{\pm0.2}$ \\
\midrule
LLM-SSC (Qwen) & 76.4$_{\pm0.3}$ & 69.7$_{\pm0.3}$ \\
SIP (Qwen) & 91.9$_{\pm0.2}$ & 84.1$_{\pm0.4}$ \\
SIP+SGVR (Qwen) & \textbf{92.9}$_{\pm0.2}$ & \textbf{84.9}$_{\pm0.3}$ \\
\bottomrule
\end{tabular}
\end{table}

Table~\ref{tab:indomain_overall} shows the results aggregated across test sets. SIP+SGVR with Qwen achieves an accuracy of 92.9 and Macro~F1 of 84.9, outperforming mBERT-HSLN by +2.3 in Macro~F1 and matching or exceeding the strongest encoder baselines, mmBERT-HSLN and XLM-R-HSLN, and clearly surpassing LaBSE-HSLN. The use of SIP and SGVR substantially improved performance over LLM-SSC across all three LLMs.

Table~\ref{tab:perlang_main} shows the results for the four major languages. Our method consistently outperformed baselines across all four languages, with the largest gains in Spanish and Japanese at +9.3 and +9.2 Macro~F1, a moderate gain of +5.8 in English, and a smaller gain of +0.9 in Chinese.

\begin{table}[t]
\centering
\small
\caption{Macro~F1 results for individual major languages. The Overall row reports the score over the pooled test sentences of all nine in-domain languages. SIP+SGVR uses Qwen2.5-3B-Instruct.}
\label{tab:perlang_main}
\begin{tabular}{lccc}
\toprule
\textbf{Language} & \textbf{mBERT} & \textbf{SIP+SGVR} & \textbf{$\Delta$} \\
\midrule
English & 84.6$_{\pm1.7}$ & 90.4$_{\pm0.0}$ & +5.8 \\
Japanese & 71.9$_{\pm2.8}$ & 81.1$_{\pm0.6}$ & +9.2 \\
Chinese & 78.2$_{\pm0.2}$ & 79.1$_{\pm0.2}$ & +0.9 \\
Spanish & 81.2$_{\pm0.8}$ & 90.5$_{\pm0.7}$ & +9.3 \\
\midrule
\textbf{Overall (9 lang.)} & \textbf{82.6}$_{\pm0.7}$ & \textbf{84.9}$_{\pm0.3}$ & \textbf{+2.3} \\
\bottomrule
\end{tabular}
\end{table}

We also analyzed component contributions, reported in Table~\ref{tab:ablation}. SGVR with a baseline prompt, namely the SIP prompt without the structural constraints and the demonstration but with the full abstract, achieved an accuracy of 91.5 and Macro~F1 of 81.6, compared to 92.9 and 84.9 for SIP+SGVR with Qwen. The $-3.3$ decrease in Macro~F1 when removing SIP indicates that the SIP prompt, which adds the ordering constraint and one demonstration to this baseline, contributes substantially to accurate classification (see Section~\ref{sec:limitations} for what this comparison does not separate), and the larger drop in Macro~F1 than in accuracy suggests an effect on minority classes. SIP+SGVR with Qwen shows improvements of +1.0 in accuracy and +0.8 in Macro~F1 over SIP alone, confirming SGVR's contribution.

\begin{table}[t]
\centering
\small
\caption{Ablation results for Qwen2.5-3B-Instruct, overall.}
\label{tab:ablation}
\begin{tabular}{lcc}
\toprule
\textbf{Method} & \textbf{Accuracy} & \textbf{Macro F1} \\
\midrule
SGVR w/o SIP & 91.5$_{\pm0.4}$ & 81.6$_{\pm0.3}$ \\
SIP only & 91.9$_{\pm0.2}$ & 84.1$_{\pm0.4}$ \\
SIP+SGVR & \textbf{92.9}$_{\pm0.2}$ & \textbf{84.9}$_{\pm0.3}$ \\
\bottomrule
\end{tabular}
\end{table}

\subsection{Zero-Shot Cross-Lingual Evaluation}
\label{subsec:zeroshot}

\begin{table}[t]
\centering
\small
\caption{Overall zero-shot evaluation results for five unseen languages.}
\label{tab:zeroshot_overall}
\begin{tabular}{lcc}
\toprule
\textbf{Method} & \textbf{Accuracy} & \textbf{Macro F1} \\
\midrule
mBERT-HSLN & 76.8$_{\pm4.3}$ & 67.7$_{\pm4.1}$ \\
mmBERT-HSLN & 80.9$_{\pm1.9}$ & 71.2$_{\pm1.7}$ \\
XLM-R-HSLN & 82.1$_{\pm1.7}$ & 72.3$_{\pm1.9}$ \\
LaBSE-HSLN & 75.6$_{\pm0.5}$ & 66.4$_{\pm0.5}$ \\
\midrule
LLM-SSC (Qwen) & 66.7$_{\pm0.3}$ & 56.8$_{\pm0.0}$ \\
SIP+SAV (Qwen) & \textbf{84.9}$_{\pm0.7}$ & \textbf{78.0}$_{\pm1.1}$ \\
\bottomrule
\end{tabular}
\end{table}

We evaluated models trained on the nine in-domain languages on five unseen languages. Since Qwen2.5-3B-Instruct was the strongest in-domain model in Table~\ref{tab:indomain_overall}, we conducted the zero-shot evaluation with Qwen alone. In this setting, validation data for the target languages were unavailable, making verifier training impossible; we therefore used SAV, which selects predictions based on candidate consistency. SIP+SAV with Qwen reaches 84.9 accuracy and 78.0 Macro~F1, outperforming the strongest encoder baseline XLM-R-HSLN at 72.3 Macro~F1 by +5.7, and mBERT-HSLN by +10.3, as reported in Table~\ref{tab:zeroshot_overall}. This indicates that encoder strength alone does not close the cross-lingual generalization gap. The larger improvement in Macro~F1 suggests more stable performance for minority classes.

Macro~F1 varied across the five unseen languages: 88.3 for Polish (30 abstracts), 87.4 for Dutch (14), 78.5 for Korean (48), 75.4 for Turkish (179), and 70.9 for Estonian (7). Because the test sets are this small, we do not interpret the differences among languages; in particular, the structural similarity of the unseen languages to the training languages cannot be estimated reliably from sets of this size and was not computed.

\section{Limitations}
\label{sec:limitations}

\textbf{Corpus-level confounding.} Each language is represented by a single collected corpus, and the corpora differ in disciplinary composition and editorial conventions and, across databases, in retrieval procedure. The structural differences reported in Section~\ref{sec:analysis} may therefore partly reflect these factors rather than conventions of the languages themselves; since differences also appear among languages collected from the same database (Chinese and Russian versus Italian and Indonesian, all from DOAJ), the database alone does not explain them.

\textbf{Subject domain bias.} Our dataset primarily comprises abstracts with explicit section headers, which are prevalent in medicine and life sciences. Its effectiveness in subject domains with different rhetorical organizations, such as humanities and social sciences, remains to be validated. A related question is whether structural similarity merely reflects domain similarity. The domain composition in Table~\ref{tab:domain_dist} differs across languages: Japanese from CiNii is almost entirely medical and life sciences, whereas Spanish and Portuguese from Dialnet include a larger share of health and social sciences. These languages nonetheless show high structural similarity and good transfer, which suggests that structural similarity is not reducible to broad domain composition alone. This is a single illustrative comparison rather than a controlled test, and the domain figures rest on partial metadata coverage, so they are approximate. The more relevant factor may be the prevalence of the IMRaD convention for structured abstracts rather than broad subject domain, and separating the two is left for future work.

\textbf{Language coverage.} Although the dataset covers 13 languages, it lacks representation for several major linguistic regions and families, including Sub-Saharan African (Swahili), Indic (Hindi), and Southeast Asian (Thai) languages. The observed structural patterns may differ for these underrepresented groups.

\textbf{Limited hyperparameter exploration.} We set $K=3$ for SGVR and $K=5$ for SAV based on preliminary experiments. The optimal value of $K$ likely varies by language, domain, and compute budget.

\textbf{Zero-shot evaluation scope.} The zero-shot evaluation was limited to five languages with small test sets of 7 to 179 abstracts, and we do not report confidence intervals for them. Scores for individual languages should therefore be read as indicative, and those from the smallest sets, Dutch and Estonian, warrant particular caution. The manual label check in Section~\ref{subsec:preprocessing} did not include these languages, so their label quality is unverified. A larger, more diverse evaluation would further strengthen our findings.

\textbf{Attribution of gains.} Table~\ref{tab:ablation} compares prompts that differ in both the ordering constraint and the demonstration, so the contribution of the constraint alone is not isolated. Likewise, the gain of SIP+SAV over LLM-SSC in Table~\ref{tab:zeroshot_overall} combines the SIP prompt, sampling of five candidates, and consistency-based selection; we did not separate these effects or compare SAV with position-wise majority voting.

\textbf{Lack of independence between language pairs.} The correlation and partial correlation analyses in Section~\ref{sec:analysis} treat the 72 language pairs as independent observations, but pairs that share a language are not statistically independent. Conventional p-values may therefore be optimistic. We did not apply a permutation or Mantel test to account for this dependence, so the reported significance levels should be interpreted with this caveat in mind. A permutation test that permutes language identities, or a regression with random effects for source and target language, would account for this dependence; we leave this to future work.

\textbf{Source language confounding.} Transfer performance correlates with source language in-domain performance, which reflects both measurable fine-tuning performance and pretraining coverage that we did not directly measure. While this confounding factor cannot be fully controlled, and is addressed in part by the partial correlations in Section~\ref{subsec:confounding}, our comparative analysis between structural similarity and linguistic proximity remains less affected, as both predictors are evaluated using the same set of language pairs under identical conditions.

\section{Conclusion}

We constructed a multilingual SSC dataset covering 13 non-English languages from five major academic databases and analyzed the factors that determine cross-lingual transfer success. Our key finding is that linguistic proximity has no consistent predictive power for transfer, whereas structural similarity in rhetorical organization, and the distribution of rhetorical roles in particular, shows a weak but consistent positive correlation. While the strength of this relationship varies across model architectures and is partially attributable to source language performance, label distribution similarity retains independent predictive power in three of four tested models.

We proposed SIP, SGVR, and SAV to leverage this structural information. In the in-domain evaluation, SIP+SGVR reaches parity with the strongest encoder baselines, mmBERT-HSLN and XLM-R-HSLN, while clearly outperforming the mBERT-HSLN and LLM-SSC baselines, and in zero-shot transfer to unseen languages SIP+SAV outperforms the strongest encoder baseline XLM-R-HSLN by +5.7 Macro~F1.

These results have practical implications for multilingual digital library systems. First, our dataset provides rhetorical annotations at the sentence level, released as full abstracts for DOAJ and HAL and as document IDs for CiNii Research, Dialnet, and TRdizin, enabling structured metadata enrichment for these collections. These annotations can support faceted search interfaces that allow users to retrieve papers by specific rhetorical components, such as finding papers with similar methods but different conclusions. Second, the correlational finding that structural similarity is associated with transfer success suggests a hypothesis for selecting training data when deploying SSC to new languages: source languages with similar rhetorical conventions may be preferable to typologically related ones. Our analysis is correlational and does not directly evaluate this selection strategy, which we leave for future work. Third, the zero-shot capability of SIP+SAV makes it feasible to extend rhetorical structure analysis to new languages in digital libraries without collecting labeled training data in each target language, lowering the barrier for multilingual deployment.

Future work includes analyzing internal representations to understand how structural patterns are encoded, extending our approach to other tasks at the discourse level such as citation function classification, verifying the structural similarity hypothesis on subsets with matched label distributions, and exploring weighting of the structural similarity components adapted to the task.

\begin{acks}
This work was supported by JSPS KAKENHI Grant Number JP25K03419. The computations were carried out on the TSUBAME4.0 supercomputer at Institute of Science Tokyo.
\end{acks}

\bibliographystyle{ACM-Reference-Format}
\bibliography{custom}

\end{document}